\documentclass{article}

\usepackage{microtype}
\usepackage{graphicx}
\usepackage{placeins}
\usepackage{subcaption}
\usepackage{booktabs} 

\usepackage{hyperref}

\usepackage[accepted]{icml2020}
\usepackage{amsmath}
\DeclareMathOperator{\concat}{\scalebox{1}[1.5]{$\parallel$}}

\icmltitlerunning{Scalable Global Attention in GCNs}

\begin{document}

\twocolumn[
\icmltitle{Permutohedral-GCN: Graph Convolutional Networks with Global Attention}



\icmlsetsymbol{equal}{*}

\begin{icmlauthorlist}
\icmlauthor{Hesham Mostafa}{Intel}
\icmlauthor{Marcel Nassar}{Intel}
\end{icmlauthorlist}

\icmlaffiliation{Intel}{Artificial Intelligence Platforms Group, Intel Corporation, San Diego, CA}

\icmlcorrespondingauthor{Hesham Mostafa}{hesham.mostafa@intel.com}

\icmlkeywords{Graph networks, Attention , Permutohedral lattice}

\vskip 0.3in
]



\printAffiliationsAndNotice{}  

\begin{abstract}
  Graph convolutional networks (GCNs) update a node's feature vector by aggregating features from its neighbors in the graph. This ignores potentially useful contributions from distant nodes. Identifying such useful distant contributions is challenging due to scalability issues (too many nodes can potentially contribute) and oversmoothing (aggregating features from too many nodes risks swamping out relevant information and may result in nodes having different labels but indistinguishable features). We introduce a global attention mechanism where a node can selectively attend to, and aggregate features from, any other node in the graph. The attention coefficients depend on the Euclidean distance between learnable node embeddings, and we show that the resulting attention-based global aggregation scheme is analogous to high-dimensional Gaussian filtering. This makes it possible to use efficient approximate Gaussian filtering techniques to implement our attention-based global aggregation scheme. By employing an approximate filtering method based on the permutohedral lattice, the time complexity of our proposed global aggregation scheme only grows linearly with the number of nodes. The resulting GCNs, which we term permutohedral-GCNs, are differentiable and trained end-to-end, and they achieve state of the art performance on several node classification benchmarks.
\end{abstract}

\section{Introduction}\label{sec:intro}
Graph convolutional networks (GCNs) and their variants achieve excellent performance in transductive node classification tasks~\cite{Defferrard_etal16,Kipf_Welling16}. GCNs use a message passing scheme where each node aggregates messages/features from its neighboring nodes in order to update its own feature vector. After $K$ message passing rounds, a node is able to aggregate information from nodes up to $K$ hops away in the graph. This message passing scheme has three shortcomings: 1) a node does not consider information from nodes that are more than $K$ hops away, 2) messages from individual nodes can get swamped by messages from other nodes making it hard to isolate important contributions from individual nodes, 3) node features can become indistinguishable as their aggregation ranges overlap, which severely degrades the performance of downstream tasks when the indistinguishable nodes belong to different classes. This over-smoothing phenomenon~\cite{Chen_etal19}) is observed even for small values of $K$ (as low as $K=3$)~\cite{Li_etal18b}

We propose a global attention-based aggregation scheme that simultaneously addresses the issues above. We modify and extend graph attention networks (GATs)~\cite{Velivckovic_etal17} to enable the attention mechanism to operate globally, i.e, a node can aggregate features from any other node in the graph weighted by an attention coefficient. This sidesteps the inherent limited aggregation range of GCNs, and allows a node to aggregate features from distant nodes without having to aggregate features from all nodes in between. Two nearby nodes in the graph can effectively aggregate information from two very different sets of nodes, alleviating the issue of over-smoothing as information does not have to come solely from their two overlapping neighborhoods.

Naive global pair-wise attention is not scalable though, as it would have to consider all pairs of nodes, and computational overhead would grow quadratically with the number of nodes. To address this, we formulate a new attention mechanism where the strength of pair-wise attention depends on the Euclidean distance between learnable node embeddings. Attention-weighted global aggregation is then analogous to a Gaussian filtering operation for which approximate techniques with linear complexity exist. In particular, we use approximate filtering based on the permutohedral lattice~\cite{Adams_etal10}. The lattice-based filtering scheme is differentiable and error backpropagates not only through the node features, but also through the node embeddings. The resulting networks, which we call permutohedral-GCNs (PH-GCNs) are thus fully differentiable. 

In PH-GCNs, node embeddings are directly learned based on the error signal of the training task (node classification task in our case). Since the attention weights between nodes increase as they move closer in the embedding space, the training error signal will move two nodes closer in the embedding space if the training loss would improve by having them integrate more information from each other. This is in contrast to prior work based on random walks~\cite{Perozzi_etal14} or matrix factorization~\cite{Belkin_Niyogi02} methods that learn embeddings in an unsupervised manner based only on the graph structure.


In our attention-weighted aggregation scheme, a node aggregates features mostly from its close neighbors in the embedding space.  Our scheme can thus be seen as a mechanism to establish soft aggregation neighborhoods in a task relevant-manner independently of the graph structure. Since graph structure is not considered in our attention-based global aggregation scheme, we concatenate node features obtained from the global aggregation scheme with node features obtained by conventional aggregation from the node's graph neighborhood. One half of the node's feature  vector is thus obtained by attention weighted aggregation from all nodes in the graph, while the other half is obtained by attention-weighted aggregation of only the nodes in the graph neighborhood. By combining the  structure-agnostic global aggregation scheme with traditional neighborhood-based aggregation, we show that PH-GCNs are able to reach state of the art performance on several node classification tasks.

\section{Related work}
In GCNs, information flows along the edges of the graph. 
The manner in which information propagates depends on the local structure of the neighborhoods: densely connected neighborhoods rapidly expand a node's influence while tree-like neighborhoods diffuse information slowly. A common way to communicate information across multiple neighborhood hubs is by constructing deeper (hierarchical) GCN architectures. 
A drawback of such GCN layer 
stacking is that it doesn't disseminate information at a uniform rate across the graph neighborhoods \cite{Xu_etal18}. Densely connected nodes with a large sphere of influence will aggregate messages  from a large number of nodes, which will over-smooth their features and make them indistinguishable from their close neighbors in the graph. On the other hand, sparsely connected nodes receive limited direct information from other nodes in the graph. Jumping knowledge networks~\cite{Xu_etal18} use skip connections to alleviate some of these problems by controlling the influence radius of each node separately. However, they still operate on a local scale which limits their effectiveness in capturing long-range node interactions. To capture contributions from distant nodes, structurally non-local aggregation approaches are needed~\cite{Wang_etal18}.



 Dynamic graph networks change the graph structure to connect semantically related nodes together without adding unnecessary depth. The node features are then updated by aggregating messages along the new edges. Dynamic graph CNNs rebuild the graph after each layer (using kd-trees) using the node features computed in the previous layer \cite{Wang_etal19}. Data-driven pooling techniques provide an alternative way to change the graph structure by clustering nodes that are semantically relevant.
These techniques will typically coarsen the graph by either selecting top scoring nodes \cite{gao2019graph}, or computing a soft assignment matrix \cite{Ying_etal18}. While rewiring the graph structure can provide long-range connections across the graph, it forces the network to learn how to retain the original graph since the original graph encodes semantically relevant relationships.

This inefficiency is addressed by graph networks that combine a traditional neighborhood aggregation scheme with a long-range aggregation scheme.  Position-aware graph neural networks achieve this by using anchor-sets. A node in the graph aggregates node feature information from these anchor sets weighted by the distance between the node and the anchor-set \cite{You19_posgnn}. Position-aware graph networks capture the local network structure as well as retain the global network position of a given node. Alternatively, Geom-GCN proposes to map the graph nodes to an embedding space via various node embedding methods \cite{Pei2020Geom-GCN}. Geom-GCN then uses the geometric relationships defined on this embedding space to build structural neighborhoods that are used in addition to the graph neighborhoods in a bi-level aggregation scheme. In contrast to our proposal, Geom-GCN uses pre-computed embeddings while we propose to learn the embeddings jointly with the graph network in an efficient and fully differentiable fashion.

Our proposed graph networks are built on top of an efficient attention mechanism. Attention was applied to graph neural networks by Veli{\v{c}}kovi{\'c} et al. \yrcite{Velivckovic_etal17}. The resulting graph attention networks (GATs) aggregate features from a node's neighbors after weighting them by attention coefficients that are produced by an attention network. The attention network is a Perceptron layer that operates on the concatenation of the features of a pair of nodes. In contrast to GAT, our attention mechanism uses Euclidean distances between learned node embeddings to compute the attention coefficients. This formulation results in an attention mechanism that is analogous to high-dimensional Gaussian filtering. Approximate Gaussian filtering methods such as the permutohedral lattice can then be used to realize scalable global attention.



The permutohedral lattice has been used in convolutional neural networks operating on sparse inputs~\cite{Su_etal18}. Permutohedral lattice convolutions have also been used to extend the standard convolutional filters so that they encompasses not only pixels in a spatial neighborhood, but also neighboring pixels in the color or intensity spaces~\cite{Jampani_etal16,Wannenwetsch_etal19}. An extension of this approach uses learned feature spaces, where the filter centered on a voxel encompasses nearby voxels in the learned feature space~\cite{Joutard_etal19}. While these approaches share the common feature of attending more strongly to nearby elements in fixed or learned feature spaces, they are not true attention mechanisms as the attention coefficients are not normalized.

\section{Methods}
\subsection{Graph convolutional networks with global attention }
We consider neural networks operating on a directed graph $G(V,E)$ where $V$ is a set of $N$ nodes/vertices and $E \subset V \times V$ is the set of edges. $v_i$ denotes the $i^{th}$ node and ${\mathcal N}(v_i)$ is the structural neighborhood of $v_i$, i.e, the indices of all nodes connected by an edge to node $v_i$.  Each node $v_i$ is associated with a feature vector ${\bf h}_i \in R^F$, where $F$ is the dimension of the input feature space. We now describe how our graph convolutional layer with global attention uses $G$ and $[{\bf h}_1,\ldots,{\bf h}_N]$ to generate for each node $v_i$ an output feature vector ${\bf h}'_i \in R^{2F'}$, where $2F'$ is the dimension of the output feature space.

The graph layer's learnable parameters are a projection matrix ${\bf W}\in R^{F' \times F}$, and a set of parameters ${\bf \Phi}$ parameterizing the generic attention function ${\mathcal A}$.  The unnormalized attention coefficient between nodes $v_i$ and $v_j$ is given by: 
\begin{equation}
  \label{eq:general_attention}
  e_{ij} = {\mathcal A}({\bf W}{\bf h}_i,{\bf W}{\bf h}_j;{\bf \Phi})
\end{equation}
To stop nodes from indiscriminately attending to all attention targets, we apply a softmax to the unnormalized attention coefficients of each node so that they sum to 1. We distinguish between structural attention where a node attends only to its graph neighbors, and global attention where a node attends to all nodes in the graph. The structural attention coefficients are given by:
\begin{equation}
\alpha^{(S)}_{ij} = \frac{exp(e_{ij})}{\sum_{k \in {\mathcal N}(v_i)}exp(e_{ik})},
\end{equation}
and they are used to aggregate features from the node's neighborhood:
\begin{equation}
{\bf h}_i'^{(S)} = \sigma\large(\sum_{j \in {\mathcal N}(v_i)} \alpha^{(S)}_{ij} {\bf W}{\bf h}_j\large),
\end{equation}
where $\sigma$ is a non-linear activation function. The global attention coefficients are given by
\begin{equation}
  \label{eq:global_coeff}
  \alpha^G_{ij} = \frac{exp(e_{ij})}{\sum_{k=1}^Nexp(e_{ik})},
\end{equation}
and they are used to aggregate features from all nodes in the graph:
\begin{equation}
  \label{eq:global_agg}
  {\bf h}_i'^{(G)} = \sigma\large(\sum_{j =1}^N \alpha^{(G)}_{ij} {\bf W}{\bf h}_j\large).
\end{equation}
We concatenate the structurally aggregated and the globally aggregated feature vectors to yield the final feature vector ${\bf h}'_i$:
\begin{equation}
{\bf h}'_i = ({\bf h}_i'^{(S)} \| {\bf h}_i'^{(G)}),
\end{equation}
where $\|$ is the concatenation operator. We denote the action of our graph convolutional layer with global attention acting in the manner described above by ${\mathcal G}(G,[{\bf h}_1,\ldots,{\bf h}_N]; {\bf W},{\bf \Phi})$. A standard technique is to use multiple attention heads~\cite{Velivckovic_etal17,Vaswani_etal17} and concatenate the outputs of the different heads. Using $A$ attention heads, the output feature vector has dimension $2AF'$ and is given by:
\begin{equation}
{\bf h}'_i = \concat_{a=1}^A {\mathcal G}(G,[{\bf h}_1,\ldots,{\bf h}_N]; {\bf W}^a,{\bf \Phi}^a).
\end{equation}

In its current form, the attention-based global aggregation scheme is hardly practical since evaluating the global attention coefficients $\alpha^G_{ij}$ and implementing the global aggregation scheme in Eq.~\ref{eq:global_agg} would need to consider all pairs of nodes in the graph. In the following section, we show that for a particular choice of the attention function ${\mathcal A}$, global aggregation (Eqs.~\ref{eq:global_coeff} and~\ref{eq:global_agg}) can be efficiently implemented using a single approximate filtering step. 

\subsection{Attention-based global aggregation and non-local filtering}
In image denoising, non-local image filters achieve superior performance compared to local filters under general statistical assumptions~\cite{Buadas_etal04}. Unlike local filters which update a pixel value based only on a spatially restricted image patch around the pixel, non-local filters average all image pixels weighted by how similar they are to the current pixel~\cite{Buades_etal05}. For $N$ points, assume point $i$ has position ${\bf p}_i \in R^D$ in the D-dim Euclidean similarity space, and an associated feature vector ${\bf f}_i \in R^F$. The similarity space may for example be the color space. A general non-local filtering operation can be written as:
\begin{equation}
{\bf f}_i' = \sum_{j=1}^N g({\bf p}_i,{\bf p}_j) {\bf f}_j,
\end{equation}
where ${\bf f}_i'$ is the output feature at position ${\bf p}_i$ and $g$ is the weighting function. The most common weighting function is the Gaussian kernel $g({\bf p}_i,{\bf p}_j) = exp(-\lambda^2||{\bf p}_i - {\bf p}_j||_2^2)$ where $||\cdot||_2$ is the 2-norm and $\lambda$ the inverse of the standard deviation. The weighting function we use is the exponential decay kernel, which yields the non-local filtering equation: 
\begin{equation}
  \label{eq:nonlocal_filtering}
  {\bf f}_i' = \sum_{j=1}^N exp(-\lambda ||{\bf p}_i - {\bf p}_j||_2) {\bf f}_j.
\end{equation}

Going back to the attention mechanism in Eq.~\ref{eq:general_attention}, GAT uses a single feedforward layer operating on the concatenated features of a pair of nodes to produce the attention coefficient between the pair. An alternative attention mechanism uses the dot product between the pair of feature vectors~\cite{Vaswani_etal17}. We introduce an attention mechanism based on Euclidean distances:
\begin{equation}
  \label{eq:euclidean_attention}
  {\mathcal A}({\bf W}{\bf h}_i,{\bf W}{\bf h}_j;{\bf \Phi}) = -\lambda||{\bf \Phi}{\bf W}{\bf h}_i - {\bf \Phi}{\bf W}{\bf h}_j||_2.
\end{equation}
${\bf \Phi} \in R^{D \times F'}$ is an embedding matrix that embeds the node's transformed features into the D-dim node similarity space. Combining equations ~\ref{eq:general_attention},~\ref{eq:global_coeff}, and~\ref{eq:global_agg}, and using the Euclidean distance form of attention in Eq.~\ref{eq:euclidean_attention}, the attention-weighted globally aggregated features at node $i$ is then:
\begin{equation}
   \label{eq:global_agg_detailed}
   {\bf h}_i'^{(G)} = \sigma\left(\frac{\sum_{j =1}^N exp\left( -\lambda||{\bf \Phi}{\bf W}{\bf h}_i - {\bf \Phi}{\bf W}{\bf h}_i  ||_2\right) {\bf W}{\bf h}_j\large}{\sum_{j =1}^N exp\left( -\lambda ||{\bf \Phi}{\bf W}{\bf h}_i - {\bf \Phi}{\bf W}{\bf h}_i ||_2 \right)}\right),
\end{equation}
which can be evaluated using two applications of the non-local filtering operation given by Eq.~\ref{eq:nonlocal_filtering}. To see that, identify:
\begin{equation}
{\bf f}_i \equiv {\bf W}{\bf h}_i \quad \text{and} \quad {\bf p}_i \equiv  {\bf \Phi}{\bf W}{\bf h}_i. 
\end{equation}
Equation~\ref{eq:global_agg_detailed} can then be written as:
\begin{equation}
  \label{eq:filtering_final}
  {\bf h}_i'^(G) = \sigma\left(\frac{\sum_{j=1}^N exp(-\lambda ||{\bf p}_i - {\bf p}_j||_2) {\bf f}_j}{\sum_{j=1}^N exp(-\lambda ||{\bf p}_i - {\bf p}_j||_2)}\right).
\end{equation}
Both the numerator and denominator are standard non-local filtering operations (all points/nodes in the denominator have a unity feature). In practice, we implement Eq.~\ref{eq:filtering_final} using only one non-local filtering step: we append 1 to the projected feature vector of every node (${\bf W}{\bf h}_j$) to get an ($F'+1$)-dim vector and then execute the non-local filtering step; we take the first $F'$ entries in the resulting feature vectors and normalize by the normalizing factor at position $F'+1$. 

\subsection{Permutohedral lattice filtering}
Exactly evaluating the non-local filtering operation in Eq.~\ref{eq:filtering_final} for all nodes still scales as $O(N^2)$ where $N$ is the number of nodes. However, approximate algorithms such as KD-trees~\cite{Adams_etal09}, and permutohedral lattice filtering ~\cite{Adams_etal10} have a more favorable scaling behavior of $O(NlogN)$ and $O(N)$, respectively. We use permutohedral lattice filtering as our approximate filtering algorithm because it is end-to-end differentiable with respect to both the node embeddings and the node features. 

\begin{figure}
  \centering
    \includegraphics[width = 0.48\textwidth]{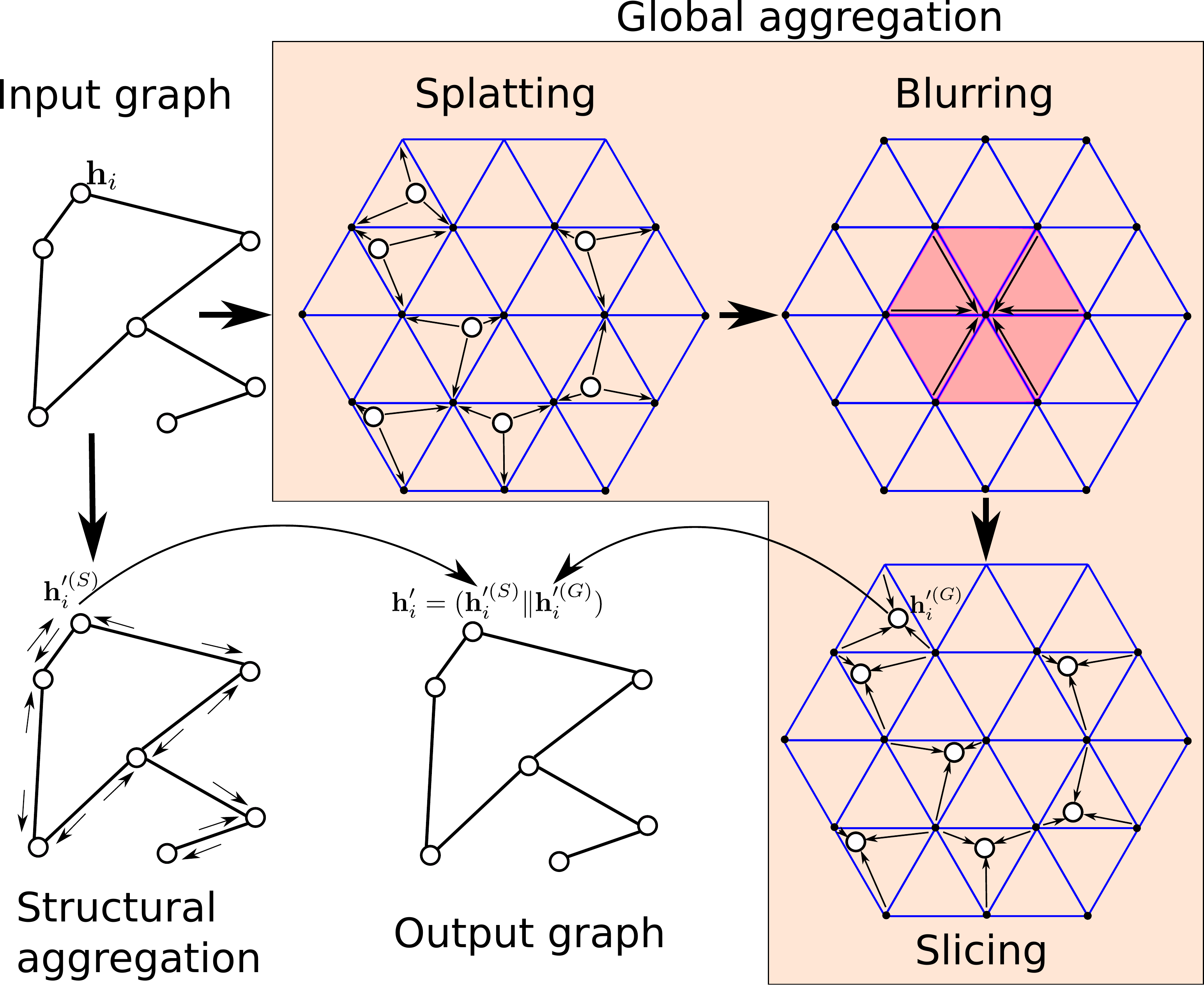} 
    \caption{ Illustration of the action of a single attention head in the PH-GCN layer. The structural aggregation pathway is similar to GAT~\cite{Velivckovic_etal17}, except that we use attention based on Euclidean distances between embeddings. Global aggregation is approximated by filtering in the permutohedral lattice which involves three steps: splatting, blurring, and slicing. The $F'$-dimensional output feature vectors from each pathway are concatenated to yield the final $2F'$-dimensional output vector. }
  \label{fig:graph_lattice}
\end{figure}

The permutohedral lattice in $(D+1)$-dim space is an integer lattice that lives within the $D$-dim hyperplane ${\bf 1}^T{\bf x} = 0$. The lattice tesselates the hyperplane with uniform simplices. There are efficient techniques for finding the enclosing simplex of any point in the hyperplane, as well as for finding the neighboring lattice points of any point in the lattice. The permutohedral lattice in the 2D plane is illustrated in Fig.~\ref{fig:graph_lattice}. Approximate filtering using the permutohedral lattice involves three steps:
\begin{itemize}
\item {\bf Splatting} : The $N$ D-dim location vectors $[{\bf p}_1,\ldots,{\bf p}_N]$ are projected onto the $D$-dim hyperplane of the permutohedral lattice. The initial feature vectors for all lattice points are initialized to zero. The lattice points defining the enclosing simplex of each projected location are found and for each location ${\bf p}_i$, the associated feature vector ${\bf f}_i$ is added to the feature vector of each enclosing lattice point after being scaled by its proximity to the lattice point. 
\item {\bf Blurring} : This step applies the filtering kernel over the lattice points. The filtering kernel decays exponentially with distance. Therefore, we calculate the output feature of a lattice point using only a small neighborhood of lattice points since contributions from more distant points would be small. This is one of the key approximations of the method: a lattice point only aggregates features from its neighborhood (weighted by the filtering kernel), instead of aggregating features from all lattice points.
\item {\bf Slicing} : After applying the approximate filtering operation over the lattice points, the output feature vector for point ${\bf p}_i$ is evaluated as the sum of the output feature vectors of its enclosing lattice points, weighted by the proximity of each enclosing lattice point to ${\bf p}_i$. 
\end{itemize}
These three steps are illustrated in Fig.~\ref{fig:graph_lattice}. Figure~\ref{fig:graph_lattice} shows the two parallel pathways used in a PH-GCN layer: the structural aggregation pathway uses attention coefficients based on the Euclidean distance between node embeddings to aggregate features from a node's immediate neighborhood; while the global aggregation pathway uses the attention coefficients to aggregate from all nodes. The $F'$-dimensional output feature vectors from each pathway are concatenated to yield the final $2F'$-dimensional output feature vector. 

\begin{figure*}[h]
  \centering
  \begin{subfigure}{0.45\textwidth}
    \includegraphics[width = \textwidth]{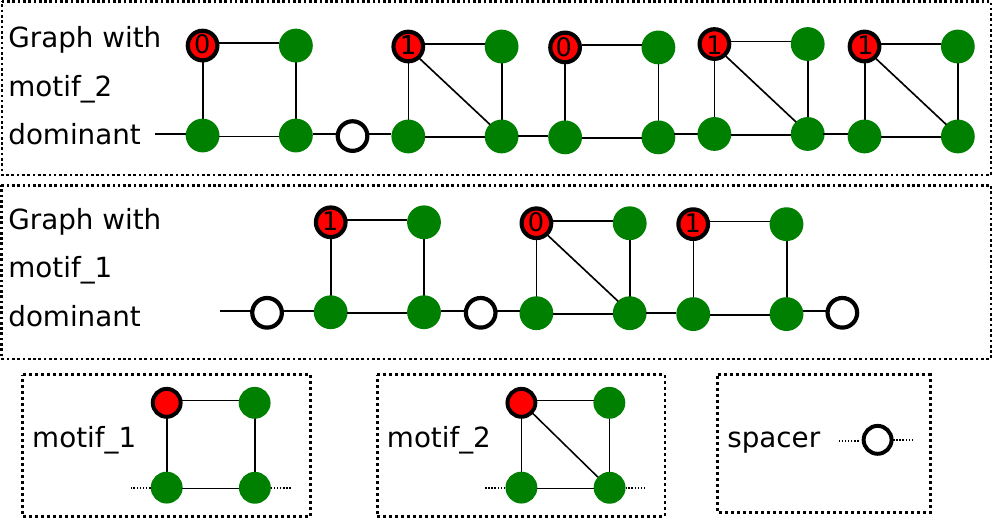} 
    \subcaption{}
    \label{fig:non_local_graph_a}
  \end{subfigure}
  \quad
  \begin{subfigure}{0.37\textwidth}
    \includegraphics[width = \textwidth]{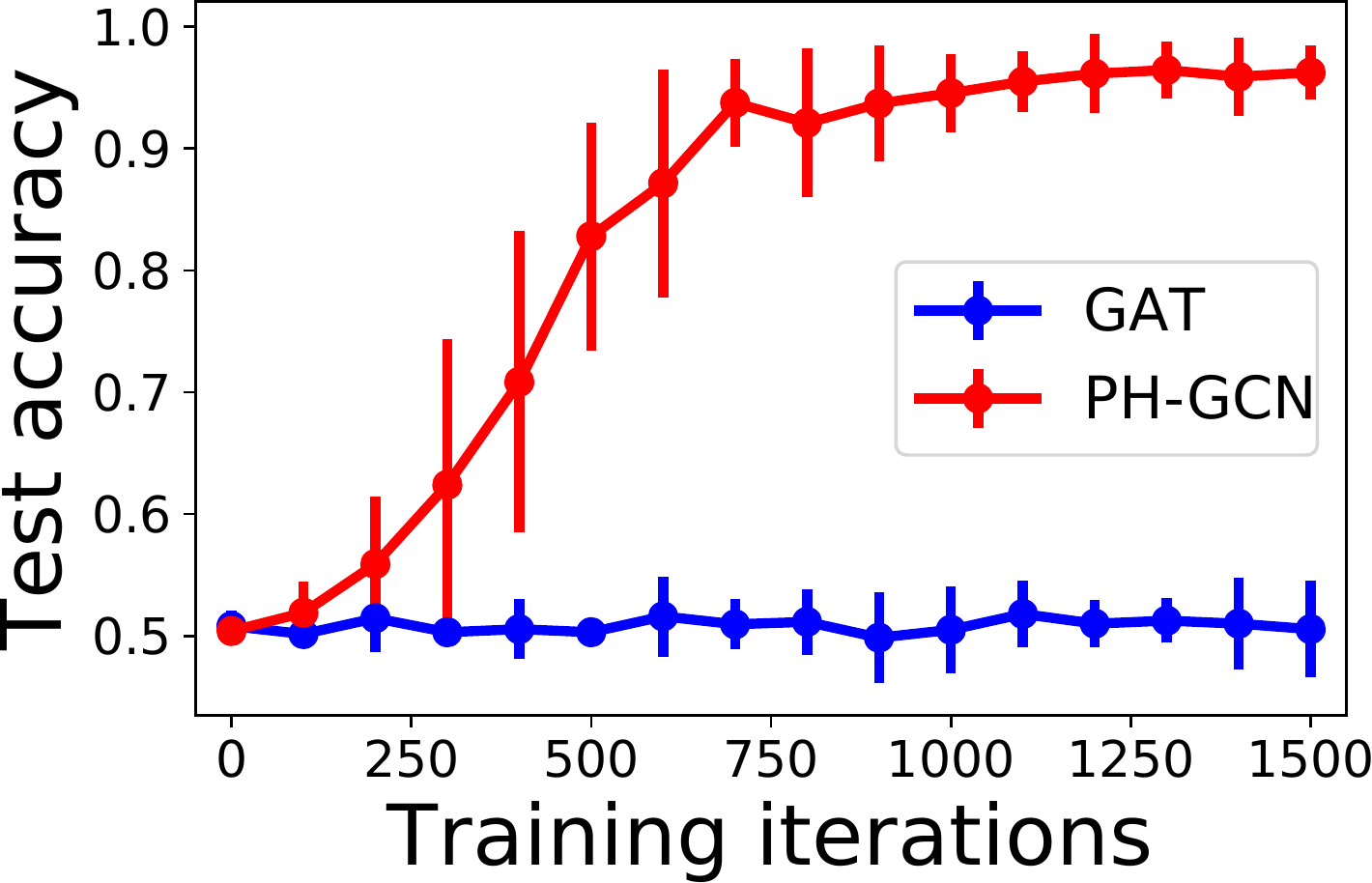} 
    \subcaption{}    
    \label{fig:non_local_graph_b}
  \end{subfigure}

  \caption{(\subref{fig:non_local_graph_a}) Two sample graphs in an inductive node classification task. The graphs have a chain structure where each element in the chain can either be motif\_1, motif\_2, or a single node (the spacer node). Each node is connected to itself (self edges not shown). Nodes with the same color have the same feature vector. The red nodes are present in motif\_1 and motif\_2. The goal is to classify each red node in the graph based on whether it is present in the dominant motif, i.e, the motif which occurs most frequently. In the middle graph for example, motif\_1 is dominant, hence red nodes in $motif\_1$ have label 1 while red nodes in motif\_2 have label 0 (the labels are shown inside the nodes). (\subref{fig:non_local_graph_b}) Accuracy of of GAT and PH-GCN as a function of training iterations. Mean and standard error bars from 10 trials.}
  \label{fig:non_local_graph}
\end{figure*}

\section{Experimental results}
\subsection{Node classification based on motif counts}
We first illustrate the power of the PH-GCN layer with its combination of structural and global aggregation on a synthetic inductive node classification task. The task is illustrated in Fig.~\ref{fig:non_local_graph}. We generate random graphs formed by a random combination of two motifs. We classify nodes in each motif based on whether the motif they are in is the motif that occurs most frequently in the graph. This is a difficult task as it requires information to flow across the whole graph in order to detect the frequency of occurrence of the different motifs and label the nodes in each motif accordingly.

We train a 3-layer GAT network and a 3-layer PH-GCN network. In every training iteration, we sample a new random chain graph composed of 10 elements (each element is either motif\_1, motif\_2, or a spacer node). We report the mean test accuracy for classifying two randomly selected nodes in 100 randomly sampled graph. As shown in Fig.~\ref{fig:non_local_graph_b}, GAT completely fails to solve the task and its performance stays at chance level, while PH-GCN is consistently able to learn the task.

The failure of GAT is expected as a 3-layer network can only aggregate node features from up to 3 hops away in the graph. This is clearly insufficient to judge whether a motif is dominant in the graph. PH-GCN on the other hand is able to learn that a node label correlates with the frequency of occurrence of its motif in the graph. This requires the neighborhood aggregation part of PH-GCN to detect the motif structure, and the global aggregation part to compare the frequency of occurrence of the two motifs. This synthetic example shows that PH-GCNs can solve problems beyond the ability of standard graph convolutional networks using the same number of layers.

\subsection{Transductive node classification}

\begin{table*}[h]
\caption{Properties of graph datasets}
\label{tab:dataset_properties}
\begin{center}
\begin{tabular}{l|ccccccc}
\hline
  & Cora & Citeseer & Pubmed & Cornell & Texas & Wisconsin & Actor \\
    \hline
    \hline
    Number of nodes & 2708 & 3327 & 19717 & 183 & 183 & 251 & 7600\\
    Number of edges & 5429 & 4732 & 44338 & 295 & 309 & 499 & 33544\\
    Node feature dimensions & 1433 & 3703 & 500 & 1703 & 1703 & 1703 & 931\\
    Number of classes & 7 & 6 & 3 & 5 & 5 & 5 & 5\\
\end{tabular}
\end{center}
\end{table*}

\begin{table*}[h]
\caption{Percentage classification accuracy. Mean and standard deviation from 10 different data splits. The Geom-GCN results were taken from the original paper.}
\label{tab:main_results}
\begin{center}
\begin{tabular}{l|ccccccc}
\hline
  & Cora & Citeseer & Pubmed & Cornell & Texas & Wisconsin & Actor \\
    \hline
    \hline
    GCN & ${\bf 88.1\pm 1.1}$ & $76.8\pm 1.2$ & $87.7\pm 0.5$ & $53.8\pm 0.8$ & $54.6\pm 1.6$ & $53.3\pm 5.2$ & $28.8\pm 0.5$\\
    GAT & $88.0\pm 1.2$ & $77.3\pm 1.1$ & $86.6\pm 0.4$ & $53.5\pm 2.3$ & $53.8\pm 2.8$ & $49.6\pm 3.4$ & $29.8\pm 1.0$\\
    Geom-GCN-I & $85.19$ & ${\bf 78.0}$ & ${\bf 90.1}$ & $56.8$ & $57.6$ & $58.2$ & $29.1$\\
    Geom-GCN-P & $84.93$ & $75.1$ & $88.1$ & $60.8$ & ${\bf 67.6}$ & $64.1$ & $31.6$\\
    Geom-GCN-S & $85.27$ & $74.7$ & $84.8$ & $55.7$ & $59.7$ & $56.7$ & $30.3$\\        
    \hline
    GAT-EDA & $87.5 \pm 1.1$ & $77.2 \pm 1.3$ & $87.1 \pm 0.2$ & $63.0\pm 3.2$ & $54.1\pm 4.4$ & $56.2\pm 3.9$ & $31.5\pm 1.1$\\
    PH-GCN & $87.9 \pm 0.7$ & $77.4 \pm 1.1$ & $89.1 \pm 0.5$ & ${\bf 74.3 \pm 6.5}$  & $63.2\pm 5.6$ & ${\bf 68.2\pm 7.3}$ & ${\bf 34.3\pm 1.3}$ \\            
\end{tabular}
\end{center}
\end{table*}

We test PH-GCN on the following transductive node classification graphs:
\begin{itemize}
\item {\bf Citation graphs} : Nodes represent papers and edges represent citations. Node features are binary 0/1 vectors indicating the absence or presence of certain key words in the paper. Each node/paper belongs to an academic topic and the goal is to predict the label/topic of the test nodes. We benchmark on three citation graphs: {\it Cora}, {\it Citeseer}, and {\it Pubmed}.
\item {\bf WebKB graphs} : The graphs capture the structure of the webpages of computer science departments in various universities. Nodes represent webpages, and edges webpage links. As in the citation graphs, node features are bag of words binary vectors. Each node is manually classed as `course', `faculty', `student', `project', or `staff'. We benchmark on three WebKB graphs: {\it Cornell}, {\it Texas}, and {\it Wisconsin}.
\item {\bf Actor co-occurrence graph} : The graph was constructed by crawling Wikipedia articles. Each node represents an actor, and an edge indicates one actor occurs on another's Wikipedia page. Node features are bag of words binary vectors. Nodes are classified into five categories based on the topic of the actor's wikipedia page. This graph is a subgraph of the film-director-actor-writer graph in ~\cite{Tang_etal09}.
\end{itemize}
Table~\ref{tab:dataset_properties} summarizes the properties of the graph datasets we use. 

We compare the performance of PH-GCN against two baselines that only use structural aggregation: GCN~\cite{Kipf_Welling16} and GAT~\cite{Velivckovic_etal17}. In addition we also test against Geom-GCN~\cite{Pei2020Geom-GCN} which uses a combination of structural aggregation and aggregation based on node proximity in an embedding space. Geom-GCN uses three different algorithms to create the node embeddings: Isomap, Poincare embeddings, and struc2vec, which leads to three different Geom-GCN variants: Geom-GCN-I, Geom-GCN-P, and Geom-GCN-S, respectively. The node embeddings used by Geom-GCN are fixed and depend only on the graph structure. Node embeddings in PH-GCN, however, are dynamic and depend on the node features. Most importantly, PH-GCN node embeddings are directly trained using the loss function of the task at hand.

PH-GCN has two novel aspects: an attention mechanism based on Euclidean distances, and an efficient scheme for attention-based global aggregation that is built on top of the new attention mechanism. We separately evaluate the performance of the new attention mechanism by replacing the attention mechanism used in GAT by our Euclidean distance attention mechanism. We term the resulting networks GAT-EDA. 

For each node classification task, we randomly split the nodes of each class into a 60\%-20\%-20\% split for training, testing, and validation respectively. During training, we monitor the validation loss and report test accuracy for the model with the smallest validation loss. We repeat all experiments 10 times for 10 different random splits and report the mean and standard deviation of the test accuracy. The hyper-parameters we tune are the number of attention heads, dropout probabilities, and the fixed learning rate. We tune these hyper-parameters to obtain best validation loss.

For PH-GCN and GAT-EDA, we always use an embedding dimension of 4. In the blurring step in the permutohedral lattice used in PH-GCN, a lattice point aggregates features from neighboring lattice points up to three hops away in all directions, i.e, we use a lattice filter of width 7. The distance scaling coefficient $\lambda$  (Eq.~\ref{eq:euclidean_attention}) was set to $1$ for structural aggregation, and set to $10$ for global aggregation. $\lambda$ scales distances in the embedding space. The larger $\lambda$ causes the attention coefficients to decay much faster with distance when doing global aggregation. This is needed to make global attention more selective as it has a much larger number of attention targets (all nodes in the graph) compared to attention in the structural aggregation step (which only considers the node's neighbors). All the networks we test have two layers. We use the ADAM optimizer~\cite{Kingma_Ba14} for all experiments. 

\begin{figure*}[h]
  \centering
  \begin{subfigure}{0.47\textwidth}
    \includegraphics[width = \textwidth]{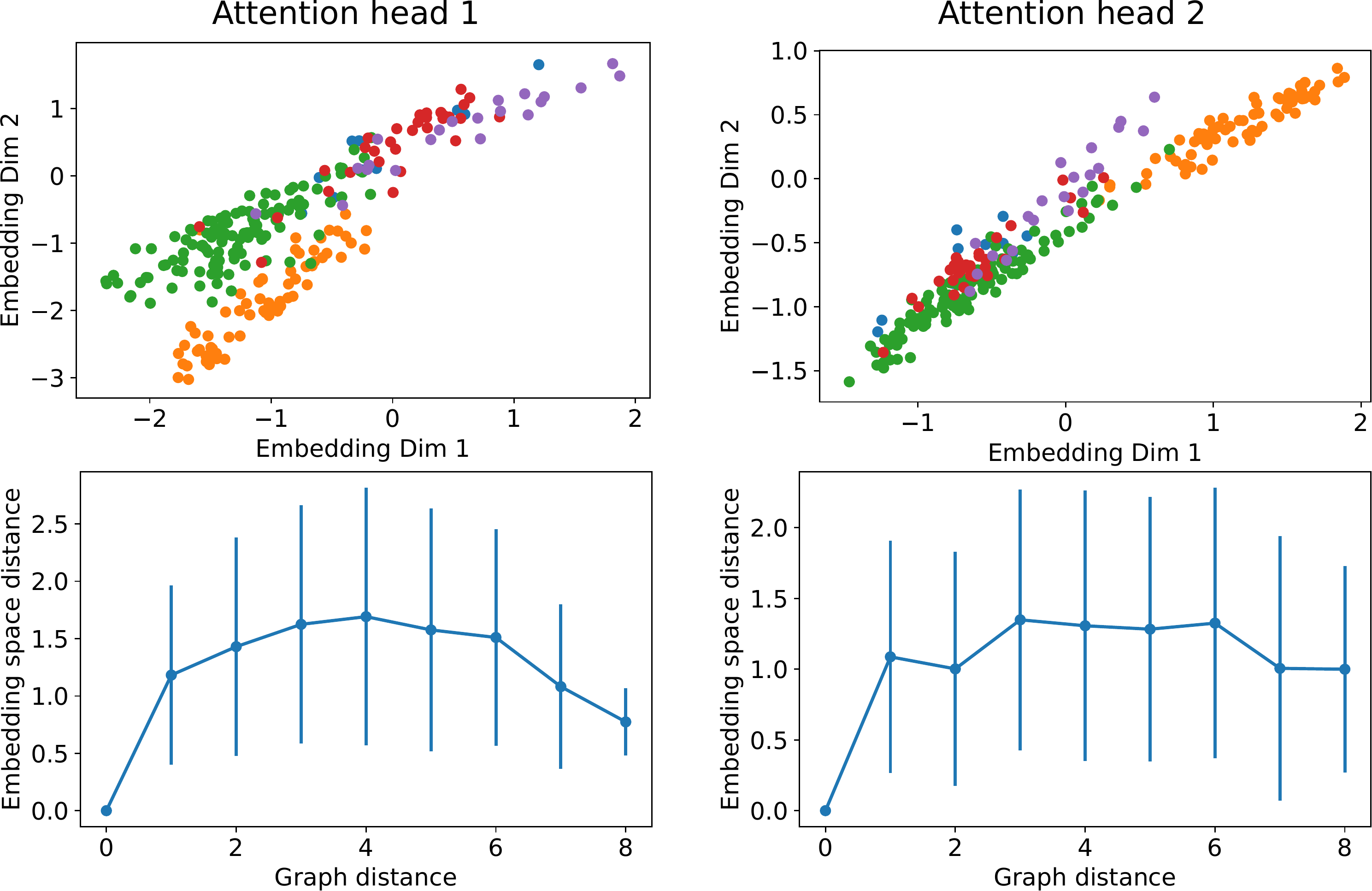} 
    \subcaption{}
    \label{fig:wisconsin_data}
  \end{subfigure}
  \quad
  \begin{subfigure}{0.47\textwidth}
    \includegraphics[width = \textwidth]{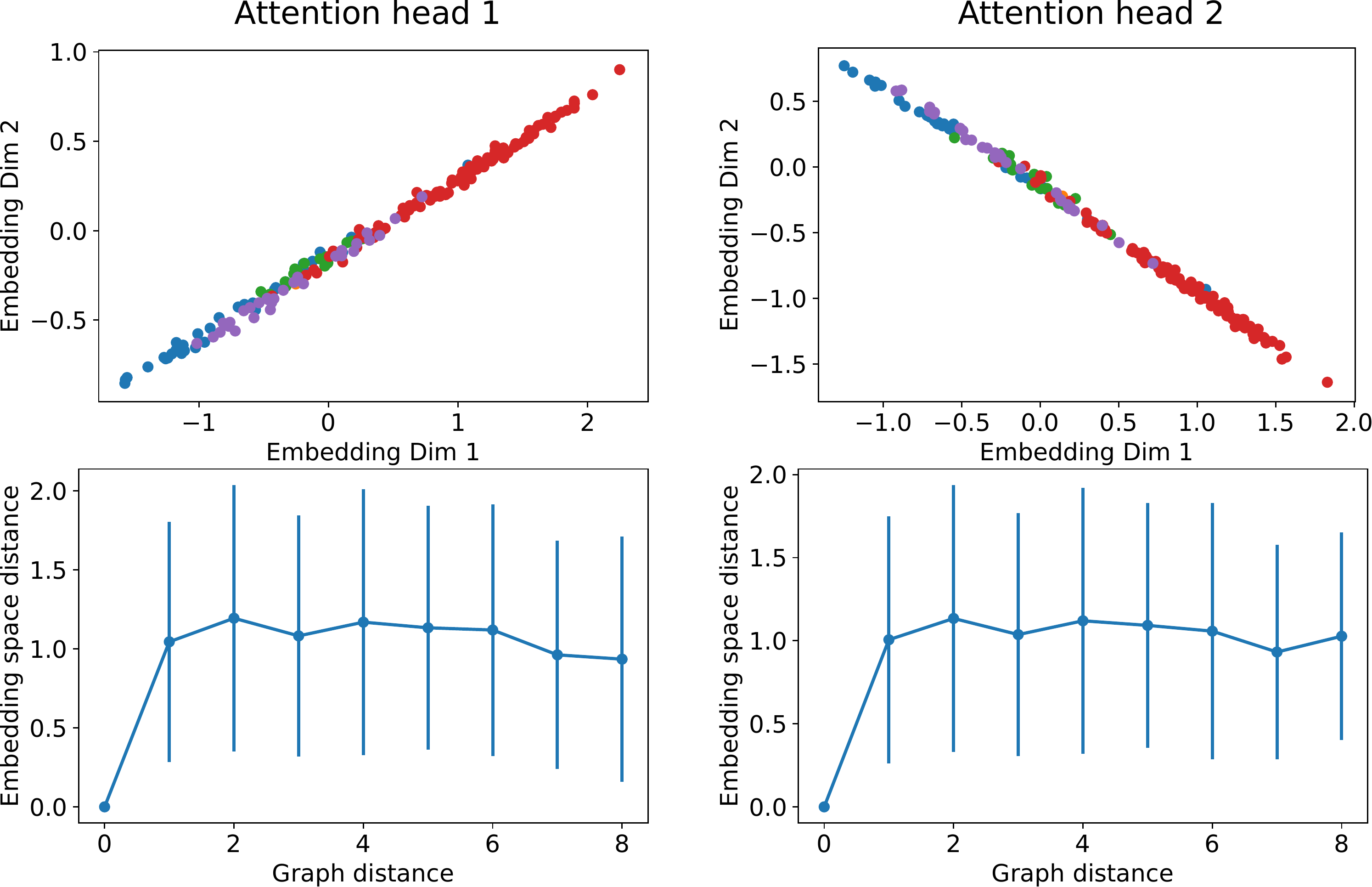} 
    \subcaption{}    
    \label{fig:cornell_data}
  \end{subfigure}

  \caption{Visualization of the learned embeddings for the two attention heads in the first PH-GCN layer when learning on  (\subref{fig:wisconsin_data}) the {\it Wisconsin} graph and (\subref{fig:cornell_data}) the {\it Cornell} graph. We used a 2D embedding space. The top row shows the embeddings of all nodes in the two graphs for the two attention heads, colored according to their class label. Nodes belonging to the same class tend to cluster together. Bottom row shows the relation between the number of hops separating two nodes (graph distance), and the distance between their learned embeddings. No clear correlation exists. }
  \label{fig:attention_viz}
\end{figure*}

The accuracy results are summarized in Table~\ref{tab:main_results}. For {\it Cora} and {\it Citeseer}, the performance of PH-GCN is not significantly different from GCN and GAT. There is a small but significant improvement of PH-GCN over GCN and GAT on the  {\it Pubmed} dataset. The lack of consistent major improvement on the three citation graphs can be attributed to their high assortativity~\cite{Pei2020Geom-GCN} where nodes belonging to the same class tend to connect together. In highly assortative graphs, a node does not need long-range aggregation in order to accumulate features from similar nodes, as similar nodes are close together in the graph anyway. Hence structural aggregation alone should perform well. Similar observations have been made regarding Geom-GCN~\cite{Pei2020Geom-GCN} which fails to show consistent improvements on the citation graphs. For the WebKB graphs and the actor co-occurrence graph, the performance advantage of PH-GCN over GCN and GAT is large and significant. For the WebKB graphs, the accuracy standard deviation can be large since the graphs are small, and differences in the training/validation/testing splits significantly alter the learning problem.

Across all graphs, PH-GCN either outperforms all Geom-GCN variants or outperforms two out of the three variants. One of the issues with Geom-GCN is that it has to pick an embedding algorithm for generating the fixed node embeddings before the start of training, and there is no clear way to select the embedding algorithm that would perform best on the task. As shown in Table~\ref{tab:main_results}, different Geom-GCN embeddings perform best on different tasks, and the accuracy difference between the different embeddings can be large. PH-GCN completely sidesteps this issue by learning the embeddings while learning the task which consistently leads to high performance.

Except for the {\it Cora} and {\it Citeseer} graphs where the performance of PH-GCN and GAT-EDA are not significantly different, PH-GCN performs significantly better than GAT-EDA. Both share the same novel Euclidean distance attention mechanism. This shows the performance advantage of PH-GCN cannot all be attributed to the new attention mechanism we use, but that global attention-based aggregation plays a crucial role.

\subsection{PH-GCN embeddings}
In this section, we investigate the properties of the node embeddings learned by PH-GCN. Each PH-GCN layer can use multiple attention heads. Each attention head learns its own node embeddings, and the Euclidean distances between these node embeddings translate to attention coefficients between the nodes. To visualize these embeddings, we use a 2D embedding space and train a 2-layer PH-GCN on the {\it Wisconsin} and {\it Cornell} graphs. We use two attention heads in the first PH-GCN layer. Figure~\ref{fig:attention_viz} shows the embeddings learned by the two heads in both tasks, colored according to their class labels. It is clear PH-GCN learns embeddings where nodes belonging to the same class are more clustered together. PH-GCN learns embeddings that minimize the classification loss; by placing nodes of the same class closer together in the embedding space, nodes belonging to one class will pre-dominantly aggregate information from each other and not from nodes belonging to other classes. Nodes belonging to the same class will thus have similar output feature vectors (since they all aggregate information from the same embedding neighborhood). In other words, by learning embeddings that cluster same-class nodes together, PH-GCN reduces the within-class variance of of the node features. This simplifies the classification task for the second (output) layer.

The embeddings learned by PH-GCN are thus directly tied to the task at hand. In Fig.~\ref{fig:attention_viz}, we plot the relation between the distance between a pair of nodes in the graph (minimum number of edges that need to be traversed to get from one node to another), and the distance between their learned embeddings. It is clear no correlation exists between the two. For all node pairs that are $K$ edges apart in the graph, there is very large variability in their separation distance in embedding space, as evidenced by the large standard deviation bars. This shows that the learned embeddings do not represent the graph structure. Instead, the embeddings represent the aggregation neighborhoods needed to perform well on the task.

\section{Discussion}
Attention-based feature aggregation is a powerful mechanism that finds use in deep learning models ranging from natural language processing (NLP) models~\cite{Vaswani_etal17} to graph neural networks~\cite{Velivckovic_etal17} to generative models~\cite{Zhang_etal18a}. It enables a model to capture relevant long-range information within a large context. Attention mechanisms, however, intrinsically have an unfavorable quadratic scaling behavior with the number of attention targets (words in an NLP setting, or nodes in a graph setting). This has traditionally been addressed by limiting the range of the attention mechanism (using small word contexts in NLP setting, or using limited node neighborhoods in graph settings).

Using approximate Gaussian filtering methods to implement global attention makes it possible to use larger attention contexts beyond what has been possible using exact attention. Our approximate formulation of global attention is directly applicable to a range of models beyond graph networks. One example is transformer models~\cite{Vaswani_etal17}. Recently, an approximate global attention mechanism based on locality sensitive hashing has been used in the transformer architecture~\cite{Kitaev_etal20} to break the quadratic scaling behavior. Unlike our approximate filtering approach, however, it has a non-differentiable step (the hashing step), and it scales as $O(NlogN)$ while our approach scales as $O(N)$ where $N$ is the number of attention targets. 

Learning informative node embeddings in graphs is a long-standing problem~\cite{Goyal_Ferrara18}. The majority of prior work, however, first chooses a similarity measure, and then finds node embeddings that put similar nodes closer together. The similarity measure could for example be adjacency in the graph~\cite{Belkin_Niyogi02} or the co-occurrence frequency of two nodes in random walks over the graph~\cite{Perozzi_etal14}. What the {\it right} similarity measure is, however, depends on the task at hand. In PH-GCN, we dispense with similarity-based node embedding approaches, and directly optimize the embeddings based on the task loss. Nodes are close together in the embedding space if the task loss improves by having them aggregate information from each other. Our approach naturally simplifies the learning problem on the graph as we do not need a separate embedding generation step.

\FloatBarrier


\begin{thebibliography}{28}
\providecommand{\natexlab}[1]{#1}
\providecommand{\url}[1]{\texttt{#1}}
\expandafter\ifx\csname urlstyle\endcsname\relax
  \providecommand{\doi}[1]{doi: #1}\else
  \providecommand{\doi}{doi: \begingroup \urlstyle{rm}\Url}\fi

\bibitem[Adams et~al.(2009)Adams, Gelfand, Dolson, and Levoy]{Adams_etal09}
Adams, A., Gelfand, N., Dolson, J., and Levoy, M.
\newblock Gaussian kd-trees for fast high-dimensional filtering.
\newblock In \emph{ACM SIGGRAPH 2009 papers}, pp.\  1--12. 2009.

\bibitem[Adams et~al.(2010)Adams, Baek, and Davis]{Adams_etal10}
Adams, A., Baek, J., and Davis, M.
\newblock Fast high-dimensional filtering using the permutohedral lattice.
\newblock In \emph{Computer Graphics Forum}, volume~29, pp.\  753--762. Wiley
  Online Library, 2010.

\bibitem[Belkin \& Niyogi(2002)Belkin and Niyogi]{Belkin_Niyogi02}
Belkin, M. and Niyogi, P.
\newblock Laplacian eigenmaps and spectral techniques for embedding and
  clustering.
\newblock In \emph{Advances in neural information processing systems}, pp.\
  585--591, 2002.

\bibitem[Buades et~al.(2004)Buades, Coll, and Morel]{Buadas_etal04}
Buades, A., Coll, B., and Morel, J.
\newblock On image denoising methods.
\newblock \emph{CMLA Preprint}, 5, 2004.

\bibitem[Buades et~al.(2005)Buades, Coll, and Morel]{Buades_etal05}
Buades, A., Coll, B., and Morel, J.
\newblock A non-local algorithm for image denoising.
\newblock In \emph{2005 IEEE Computer Society Conference on Computer Vision and
  Pattern Recognition (CVPR'05)}, volume~2, pp.\  60--65. IEEE, 2005.

\bibitem[Chen et~al.(2019)Chen, Lin, Li, Li, Zhou, and Sun]{Chen_etal19}
Chen, D., Lin, Y., Li, W., Li, P., Zhou, J., and Sun, X.
\newblock Measuring and relieving the over-smoothing problem for graph neural
  networks from the topological view.
\newblock \emph{arXiv preprint arXiv:1909.03211}, 2019.

\bibitem[Defferrard et~al.(2016)Defferrard, Bresson, and
  Vandergheynst]{Defferrard_etal16}
Defferrard, M., Bresson, X., and Vandergheynst, P.
\newblock Convolutional neural networks on graphs with fast localized spectral
  filtering.
\newblock In \emph{Advances in neural information processing systems}, pp.\
  3844--3852, 2016.

\bibitem[Gao \& Ji(2019)Gao and Ji]{gao2019graph}
Gao, H. and Ji, S.
\newblock Graph {U-nets}.
\newblock In \emph{Proceedings of The 36th International Conference on Machine
  Learning}, 2019.

\bibitem[Goyal \& Ferrara(2018)Goyal and Ferrara]{Goyal_Ferrara18}
Goyal, P. and Ferrara, E.
\newblock Graph embedding techniques, applications, and performance: A survey.
\newblock \emph{Knowledge-Based Systems}, 151:\penalty0 78--94, 2018.

\bibitem[Jampani et~al.(2016)Jampani, Kiefel, and Gehler]{Jampani_etal16}
Jampani, V., Kiefel, M., and Gehler, P.
\newblock Learning sparse high dimensional filters: Image filtering, dense crfs
  and bilateral neural networks.
\newblock In \emph{Proceedings of the IEEE Conference on Computer Vision and
  Pattern Recognition}, pp.\  4452--4461, 2016.

\bibitem[Joutard et~al.(2019)Joutard, Dorent, Isaac, Ourselin, Vercauteren, and
  Modat]{Joutard_etal19}
Joutard, S., Dorent, R., Isaac, A., Ourselin, S., Vercauteren, T., and Modat,
  M.
\newblock Permutohedral attention module for efficient non-local neural
  networks.
\newblock In \emph{International Conference on Medical Image Computing and
  Computer-Assisted Intervention}, pp.\  393--401. Springer, 2019.

\bibitem[Kingma \& Ba(2014)Kingma and Ba]{Kingma_Ba14}
Kingma, D. and Ba, J.
\newblock Adam: A method for stochastic optimization.
\newblock \emph{arXiv preprint arXiv:1412.6980}, 2014.

\bibitem[Kipf \& Welling(2016)Kipf and Welling]{Kipf_Welling16}
Kipf, T. and Welling, M.
\newblock Semi-supervised classification with graph convolutional networks.
\newblock \emph{arXiv preprint arXiv:1609.02907}, 2016.

\bibitem[Kitaev et~al.(2020)Kitaev, Kaiser, and Levskaya]{Kitaev_etal20}
Kitaev, N., Kaiser, {\L}., and Levskaya, A.
\newblock Reformer: The efficient transformer.
\newblock \emph{arXiv preprint arXiv:2001.04451}, 2020.

\bibitem[Li et~al.(2018)Li, Han, and Wu]{Li_etal18b}
Li, Q., Han, Z., and Wu, X.-M.
\newblock Deeper insights into graph convolutional networks for semi-supervised
  learning.
\newblock In \emph{Thirty-Second AAAI Conference on Artificial Intelligence},
  2018.

\bibitem[Pei et~al.(2020)Pei, Wei, Chang, Lei, and Yang]{Pei2020Geom-GCN}
Pei, H., Wei, B., Chang, K. C.-C., Lei, Y., and Yang, B.
\newblock Geom-gcn: Geometric graph convolutional networks.
\newblock In \emph{International Conference on Learning Representations}, 2020.
\newblock URL \url{https://openreview.net/forum?id=S1e2agrFvS}.

\bibitem[Perozzi et~al.(2014)Perozzi, Al-Rfou, and Skiena]{Perozzi_etal14}
Perozzi, B., Al-Rfou, R., and Skiena, S.
\newblock Deepwalk: Online learning of social representations.
\newblock In \emph{Proceedings of the 20th ACM SIGKDD international conference
  on Knowledge discovery and data mining}, pp.\  701--710, 2014.

\bibitem[Su et~al.(2018)Su, Jampani, Sun, Maji, Kalogerakis, Yang, and
  Kautz]{Su_etal18}
Su, H., Jampani, V., Sun, D., Maji, S., Kalogerakis, E., Yang, M.-H., and
  Kautz, J.
\newblock Splatnet: Sparse lattice networks for point cloud processing.
\newblock In \emph{Proceedings of the IEEE Conference on Computer Vision and
  Pattern Recognition}, pp.\  2530--2539, 2018.

\bibitem[Tang et~al.(2009)Tang, Sun, Wang, and Yang]{Tang_etal09}
Tang, J., Sun, J., Wang, C., and Yang, Z.
\newblock Social influence analysis in large-scale networks.
\newblock In \emph{Proceedings of the 15th ACM SIGKDD international conference
  on Knowledge discovery and data mining}, pp.\  807--816, 2009.

\bibitem[Vaswani et~al.(2017)Vaswani, Shazeer, Parmar, Uszkoreit, Jones, Gomez,
  Kaiser, and Polosukhin]{Vaswani_etal17}
Vaswani, A., Shazeer, N., Parmar, N., Uszkoreit, J., Jones, L., Gomez, A.,
  Kaiser, {\L}., and Polosukhin, I.
\newblock Attention is all you need.
\newblock In \emph{Advances in neural information processing systems}, pp.\
  5998--6008, 2017.

\bibitem[Veli{\v{c}}kovi{\'c} et~al.(2017)Veli{\v{c}}kovi{\'c}, Cucurull,
  Casanova, Romero, Lio, and Bengio]{Velivckovic_etal17}
Veli{\v{c}}kovi{\'c}, P., Cucurull, G., Casanova, A., Romero, A., Lio, P., and
  Bengio, Y.
\newblock Graph attention networks.
\newblock \emph{arXiv preprint arXiv:1710.10903}, 2017.

\bibitem[Wang et~al.(2018)Wang, Girshick, Gupta, and He]{Wang_etal18}
Wang, X., Girshick, R., Gupta, A., and He, K.
\newblock Non-local neural networks.
\newblock In \emph{Proceedings of the IEEE conference on computer vision and
  pattern recognition}, pp.\  7794--7803, 2018.

\bibitem[Wang et~al.(2019)Wang, Sun, Liu, Sarma, and Bronstein]{Wang_etal19}
Wang, Y., Sun, Y., Liu, Z., Sarma, S.~E., and Bronstein, M.~M.
\newblock Dynamic graph cnn for learning on point clouds.
\newblock \emph{ACM Transactions on Graphics (TOG)}, 38\penalty0 (5), 2019.

\bibitem[Wannenwetsch et~al.(2019)Wannenwetsch, Kiefel, Gehler, and
  Roth]{Wannenwetsch_etal19}
Wannenwetsch, A., Kiefel, M., Gehler, P., and Roth, S.
\newblock Learning task-specific generalized convolutions in the permutohedral
  lattice.
\newblock In \emph{German Conference on Pattern Recognition}, pp.\  345--359.
  Springer, 2019.

\bibitem[Xu et~al.(2018)Xu, Li, Tian, Sonobe, ichi Kawarabayashi, and
  Jegelka]{Xu_etal18}
Xu, K., Li, C., Tian, Y., Sonobe, T., ichi Kawarabayashi, K., and Jegelka, S.
\newblock Representation learning on graphs with jumping knowledge networks.
\newblock \emph{arXiv preprint arXiv:1806.03536}, 2018.

\bibitem[Ying et~al.(2018)Ying, You, Morris, Ren, Hamilton, and
  Leskovec]{Ying_etal18}
Ying, Z., You, J., Morris, C., Ren, X., Hamilton, W., and Leskovec, J.
\newblock Hierarchical graph representation learning with differentiable
  pooling.
\newblock In \emph{Advances in neural information processing systems}, pp.\
  4800--4810, 2018.

\bibitem[You et~al.(2019)You, Ying, and Leskovec]{You19_posgnn}
You, J., Ying, R., and Leskovec, J.
\newblock Position-aware graph neural networks.
\newblock In \emph{Proceedings of The 36th International Conference on Machine
  Learning}, 2019.

\bibitem[Zhang et~al.(2018)Zhang, Goodfellow, Metaxas, and
  Odena]{Zhang_etal18a}
Zhang, H., Goodfellow, I., Metaxas, D., and Odena, A.
\newblock Self-attention generative adversarial networks.
\newblock \emph{arXiv preprint arXiv:1805.08318}, 2018.

\end{thebibliography}
\end{document}